# Causal Inference for Banking, Finance, and Insurance – A Survey

Satyam Kumar[1], Yelleti Vivek[1], Vadlamani Ravi[1,*], Indranil Bose[2]

[1]Center for Artificial Intelligence and Machine Learning,
Institute for Development and Research in Banking Technology (IDRBT),
Castle Hills Road #1, Masab Tank, Hyderabad-500076, India.
[2] Indian Institute of Management Ahmedabad,
Vastrapur, Ahmedabad 380015, India.

learnsatyam@gmail.com; yvivek@idrbt.ac.in; vravi@idrbt.ac.in; indranilb@iima.ac.in

**Abstract**

Causal Inference plays an significant role in explaining the decisions taken by statistical models and artificial intelligence models. Of late, this field started attracting the attention of researchers and practitioners alike. This paper presents a comprehensive survey of 37 papers published during 1992-2023 and concerning the application of causal inference to banking, finance, and insurance. The papers are categorized according to the following families of domains: (i) Banking, (ii) Finance and its subdomains such as corporate finance, governance finance including financial risk and financial policy, financial economics, and Behavioral finance, and (iii) Insurance. Further, the paper covers the primary ingredients of causal inference namely, statistical methods such as Bayesian Causal Network, Granger Causality and jargon used thereof such as counterfactuals. The review also recommends some important directions for future research. In conclusion, we observed that the application of causal inference in the banking and insurance sectors is still in its infancy, and thus more research is possible to turn it into a viable method.

---

# 1. Introduction

In fields like behavioral science, research is driven by causal rather than statistical inferences obtained from historical data. One of the frequently asked questions in any company or organization is, *"Can data alone demonstrate the company reputation and its engagement in hiring discrimination?"*. Answering this kind of queries requires some knowledge from data-generating process, and regardless of the sample size, one cannot accomplish this solely based on the given data (Pearl, 2010).

Causal Inference refers to the process of determining the causal relationship between variables or events. It involves establishing a cause-and-effect connection by analyzing data and drawing conclusions about the influence one variable has on another. Causal Inference is a fundamental concept in statistics, epidemiology, and other disciplines that aims to understand cause-and-effect relationships between variables. It refers to the process of making inferences about causal relationships based on observational or experimental data. The genesis of causal inference can be traced back to the works of philosophers and statisticians who recognized the need to establish causality rather than mere correlation in scientific investigations.

In various fields, causal inference offers numerous benefits. In epidemiology, it enables researchers to determine the effectiveness of interventions and treatments by identifying the causal relationship between an exposure (e.g., a drug or environmental factor) and an outcome (e.g., a disease or health condition) (Pearl, 2009).

Furthermore, in the field of economics, causal inference allows economists to evaluate the impact of specific factors on economic outcomes, such as the effect of minimum wage laws on employment levels or the influence of tax policies on economic growth (Angrist & Pischke, 2009). It helps researchers and policymakers in decision-making processes by providing evidence-based insights into the potential consequences of different actions.

Certain extensions to the standard statistical mathematical language are required for solving systematic causal problems. The Causal Hierarchy Theorem (CHT) states that the three layers of the hierarchy are typically separated in a measure-theoretic sense, meaning that data from one layer does not fully determine information at higher layers. As scientists often lack access to the precise causal mechanisms, inferences are studied using graphical methods. More precisely, it helps us understand who we are, how to relate with others, and how to make sense of our surroundings (Bareinboim et al., 2022). Causal information holds great value and is sought after in various areas of human pursuits such as science, engineering, business, and law. The concept of causality is a critical aspect of our investigation of the physical world and our comprehension of nature. The accepted scientific approach involves gathering information through observations and experimenting with various causes depending on the use case, thereby utilizing it to construct theories about undiscovered causal processes (Bareinboim et al., 2022).

Over time, there has been much deliberation on the idea of causality, yet it remains an important form of knowledge due to its ability to provide guidance on how to achieve a desired outcome while avoiding unfavorable results. Causality involves connections in which an alteration in one factor produces a corresponding modification in another factor. This relationship is dependent on three factors: correlation, temporal order, and control of "third variables" - alternative explanations for the displayed causal relationship. It's crucial to consider the possibility of spurious relationships, where the correlation between variables appears to indicate a causal effect, but instead effectuates from a hidden shared cause. In essence, causality addresses connections among variables in which a change in one factor will inevitably trigger a change in another (Oppewal, 2010).

Causal Inference methods are crucial in evaluating the effectiveness of financial regulations. Researchers analyze the impact of regulatory changes on various outcomes, such as bank stability, risk-taking behavior, or market liquidity. By employing techniques such as regression discontinuity design or synthetic control methods, they can identify the causal effect of specific regulations. For instance, Demirguc-Kunt & Huizinga (2009) had hinted that a study to examine the causal relationship between bank activities and risk-taking behavior, providing valuable insights for policymakers and regulators can be conducted in near future.

Traditional ML models focus on associations rather than causation. Explainable AI deals with the implementation of transparency and traceability of black-box statistical and machine learning methods, particularly deep learning (DL). We argue that there is a need to go beyond explainable AI. Interpretable machine learning has seen an increase in the number of causality-oriented algorithms as the popularity of causal inference has grown. In contrast to conventional methods, causal approaches are used to analyze model design's choices and actions or to determine its cause and effect (Xu et al.,2020). Traditional interpretable models are unable to provide critical insights into machine learning models. Traditional interpretability frameworks, for example, are unable to respond to causal questions, such as "*What is the impact of the $n^{th}$ filter of the $m^{th}$ layer of a deep neural network on the predictions of the model?*" that are important and necessary for comprehending a neural network model (Moraffah et al.,2020).

Kim & Bastani, (2019), proposed a framework to convert any algorithm into an interpretable individual treatment effect estimation framework, bridging the gap between causal and interpretable models. The three task categories of description, prediction, and counterfactual prediction can be used to categorize the scientific contributions of data science. Data is used in description to give a quantitative breakdown of some aspects of the world. Prediction is the mapping of certain world features (the inputs) to other world features using data (the outputs). The analytical techniques used for prediction range from simple computations (such as a correlation coefficient or risk difference) to complex pattern recognition techniques and supervised learning algorithms that can be utilized as classifiers. In counterfactual prediction, specific aspects of the world are predicted using data as if the world had been different. For all three tasks, statistical inference is frequently necessary. However, this category is the starting point for effectively completing each data science activity i.e., data requirements, presumptions, and analytics.

## 1.1 Theme of the survey

This paper presents a comprehensive review of the works reported during 1993-2023 in causal inference as applied to banking, finance, and insurance problems. We collected 37 papers falling in this area.

Fig. 1 depicts the selection process of causal inference articles chosen for review. All these papers were collected from publication sources which include a list of journals as mentioned in Table 3, list of conferences as specified in Table 4 and remaining article details are provided in Table 5. Once the publication sources were finalized, then further filtration of articles was effected by using the following keywords namely, (i) causal inferences, (ii) counterfactuals, (iii) Bayesian causal network, (iv) Granger causality, and (v) sometimes other approaches presented in causal inference papers such as regression discontinuity, difference-in-difference, structural equation model (SEM) to filter out articles. Furthermore, filter options in publication sources for selecting journals, conference papers and book chapters were utilized. Final step was filtering out papers related to the domain which includes papers from overlapping domain such as financial economics, behavioral finance, corporate governance and assets, investments based on abstract and keywords from the recommended list of papers. Most of the times recommended list featured papers from other domains such as in econometrics having following content household income, Gross domestic product (GDP), impact of military taxation, business, and marketing areas are excluded from the survey.

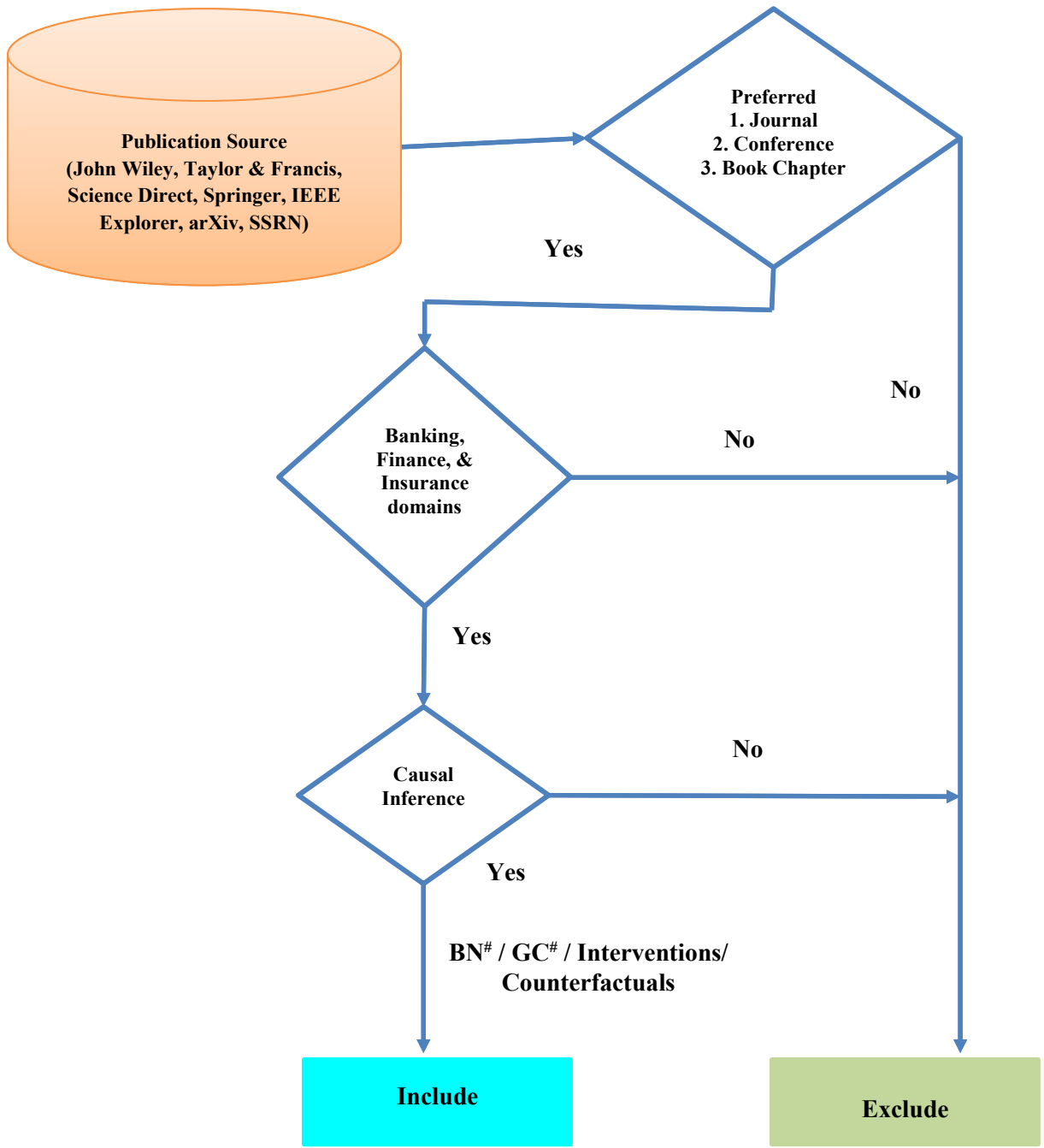

**Fig. 1.** Methodology followed while choosing the articles for review.

# BN: Bayesian Network, # GC: Granger Causality

**Table 1:** Topics obtained after performing topic-modeling on the abstracts

| Topic ID | Top 10 Topic words |
|---|---|
| Topic -1 | ['model', 'counterfactual', 'market', 'causal', 'risk', 'algorithm', 'portfolio', 'study', 'data', 'result'] |
| Topic -2 | ['financial', 'causal', 'network', 'rate', 'risk', 'framework', 'causality', 'change', 'price', 'model'] |
| Topic -3 | ['causal', 'model', 'data', 'effect', 'paper', 'inference', 'study', 'loan', 'infer', 'accounting'] |
| Topic -4 | ['contagion', 'result', 'decision', 'estimation', 'future', 'bayesian', 'credit', 'counterfactual', 'market', 'central'] |
| Topic -5 | ['causal', 'banking', 'system', 'inference', 'customer', 'research', 'study', 'rare', 'data', 'churn'] |

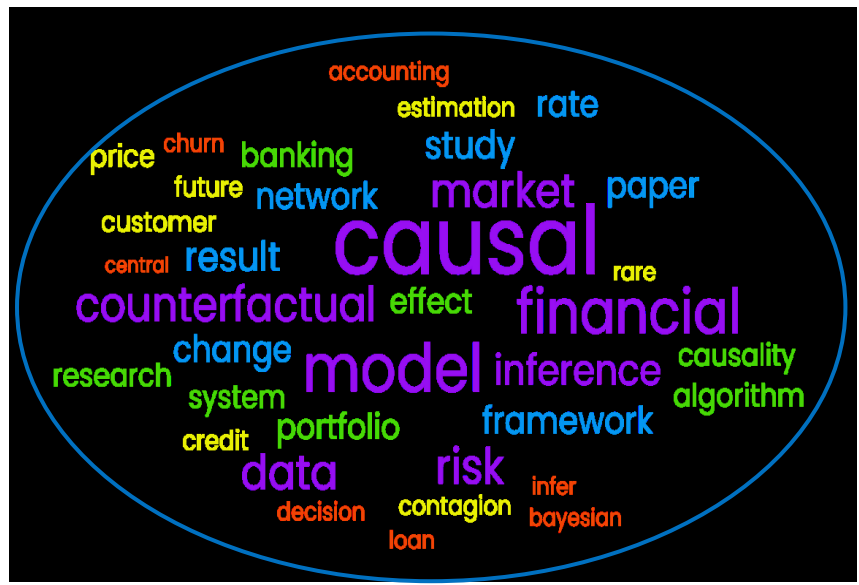

**Fig. 2.** Word cloud of the most frequent topic words of the abstracts of all the surveyed articles

To present a word cloud of the primary key words in the abstracts of the surveyed papers, we performed topic-modeling on the abstracts by using Latent-Dirichlet analysis (LDA) (Blei et al, 2003). Then, we presented the themes of the respective topic and the corresponding top 10 most repeated words within the topic in Table 1 and Fig.2. The most often repeated words are causal, inference, and modeling across the topics. This word cloud is generated using the freely available online tool[1].

## 1.2 Existing surveys on Causal Inference

When the experiments are not conducted in a controlled environment, then the cause-effect study becomes too complex. This is mainly evident in empirical archival financial accounting. Gassen, (2014) presented the various studies published during the period 2000-2012 in various fields such as Accounting, Finance, Economics, Management and Marketing. He also discussed the underlying challenges. He proposed a five-stage methodology which focuses on developing a quasi-experimental setup to mitigate the challenges in handling archival financial accounting.

---
[1] https://www.freewordcloudgenerator.com/generatewordcloud

Atansov & Black, (2016) conducted a survey on the shock-based methods for causal inference in finance and accounting research field. The authors reported 74 different methods in their survey paper. This survey primarily focuses on the studies where the impact of changes in governance over firms' approach is studied. The shock is defined as an external discrete event which impacts the firms either as a treatment or control. To handle these issues, various shock-based designs such as difference in differences (DiD), event studies (ES), instrumental variables (IV) etc., were proposed handling the selection bias.

Gow et al., (2016) conducted a study to show the use of the causal inference in accounting research and draw its inference based on observational data. The authors surveyed 105 papers related to field data and archival data which were published in Journal of Accounting Research, The accounting review and Journal of Economics and Accounting during 2014. The authors discussed the limitations of applying quasi experimental setup in accounting research. Further, they discussed the importance of structural models in identifying the causal pathways and the limitations of their application in accounting research.

Delis et al., (2020) presents a model of management practices as an unobserved (latent) input in a standard production function of banks. The authors derive estimates of management at the bank-quarter level, using data for all U.S. banks from 1984 to 2016 and Bayesian techniques. The Bayesian approach to estimate management practices allows them to incorporate prior information about management quality into their estimates. The result has shown that that management practices are highly correlated with bank profitability and default risk. This suggests that management quality is an important determinant of bank performance.

Fukuyama et al. (2023) proposes a dynamic network data envelopment analysis (DNDEA) behavioral model with a sequential structure to measure the efficiency of Chinese commercial banks. The model incorporates the dual-role characteristics of production factors, which means that a factor can be both an input and an output. The model is also validated using causal analysis. The banks were more efficient in 2010 than in 2018. The authors conclude that the DNDEA behavioural model with a sequential structure is a useful tool for measuring the efficiency of Chinese commercial banks. The model can be used to identify inefficient banks and to track their performance over time. The model can also be used to help banks improve their efficiency.

The current survey is distinct from the existing survey papers as follows:
- Extant survey papers focused on the impact of causal inference on finance only, especially accounting. However, the current survey paper includes the applications of causal inference on all finance-related domains such as accounting, corporate finance, corporate governance, explainability in financial services such as credit lending, credit rating, and in churn modelling, insurance and importantly in banking. Thus, our survey is more comprehensive covering all financial services.
- Further, we proposed a modified CRISP-DM, by incorporating causal inference and XAI. The need for the hour is also discussed in detail.

The following are the research questions addressed in this survey:
- Discussing the need to employ causal inference and explainability in the current analytical customer relationship management framework in the financial services industry in order to make it more customer-centric and responsible.
- Studying the impact of causal inference and explainability from the business perspective in understanding the deployed models.
- Development of transparent and accountable-AI systems by employing the causal inference methods, and performing 'what-if' analysis.

This paper is organized as follows: Section 2 provides an overview of causal inference. This section lays the foundation of causal inference from association to causal analysis (Section 2.2), representation of causal analysis by directed acyclic graphs (refer to Section 2.3), and later providing explanations for counterfactuals (refer to Section 2.4) and causal effect estimation (refer to Section 2.5). Section 3 describes the statistical methods used for determining causal inference, such as Bayesian Networks and Granger Causality which is a popular causal method for time series. Section 4 discusses the existing work carried out in CRISP-ML and our extension to CRISP-DM. Section 5 provides a survey of the literature carried out in the area of BFSI. Section 6 provides a list of software and tools essential for causal inference which will be beneficial for researchers in this niche area. Section 7 provides some implications for managers and researchers that are going to use this survey as a reference. Section 8 leads to the conclusion of this survey, and Section 9 discusses some of the future directions for the researchers who wish to work in this niche area of employing causal inference in BFSI.

# 2. Overview of Causal Inference

This section provides overview of the foundation which includes assumptions and notation that are required for understanding causal inference.

## 2.1 Correlation Does Not Imply Causation

Reichenbach (1956) introduced the common cause principle which mentions three possibilities. According to this principle, probabilistic correlations between events and causal structure are linked if two random X and Z are statistically dependent (refer to Fig.3) then (a) X causes Z, (b) Z causes X, and (c) there exists third variable Y that causes both X and Y. X and Z are then said to be conditional independent of Y

Computer scientist Pearl (Pearl,2010) developed the Ladder of Causality, a framework that focuses on the different roles of seeing, doing, and imagining, which has led to advances in understanding causality. It is called Pearl's Causal Hierarchy (PCH) and has a three-level causal model, namely association, intervention, and counterfactuals (or hypothetical). Each level deals with different types of questions that require a basic understanding of lower levels to answer higher level questions (Bareinboim et al., 2022). Indeed, we want to answer questions about intervention and type of association before answering retrospective queries. PCH Level 1 focuses on observations and information. For the effects of actions, Level 2 encodes information about what might happen if the intervention were to occur. Finally, Layer 3 answers hypothetical questions regarding what would have happened if a certain intervention had been made when something else had happened.

## 2.2 From Association to Causal Analysis

Any relationship that can be described in terms of a joint distribution of observed variables is an idea of association and any relationship that cannot be described solely from the distribution is a causal concept. Correlation, regression, dependency, conditional independence, probability, collapse, propensity score, risk ratio, odds ratio, marginalization, Granger causality, conditionalization, *"controlling for,"* and so on are examples of associational notions. Causation fundamentals include randomization, impact, effect, confounding, "*holding constant*," disturbance, error terms, structural coefficients, spurious correlation, faithfulness/stability, instrumental variables, intervention, explanation, and attribution (Pearl, 2010).

Causal analysis must adopt new notation for expressing causal interactions as causal relationships cannot be expressed as a joint conditional probability. Causal analysis takes a step further by attempting to deduce not just the likelihood of events under stable settings but also the dynamics of events under changing conditions, such as those brought on by medical treatments or outside interventions (Pearl, 2011).

## 2. 3 Directed Acyclic Graphs

Statistical dependence between three variables A, B, C can be visualized by a directed acyclic graph (DAG). DAG, G, is the tuple consisting of a set of vertices $X= \{X_1, X_2…, X_n\}$ and a set of edges E represented as G= (X, E) with no cycles. If there is a directed edge between 2 nodes, for instance A→B that represents a direct conditional relationship between A and B or under a causal assumption, means that A is a direct cause of B.

There exist three possibilities of DAGs to represent X and Z are marginally dependent but conditionally independent given Y as depicted in Fig. 3. In short, it can be said that X and Z are dependent nodes and dependent on node Y, which blocks the path between X and Z. Y is called collider. Conditioning on a collider and then computing the association between X and Z will lead to a different estimate, and the induced bias is known as collider bias.

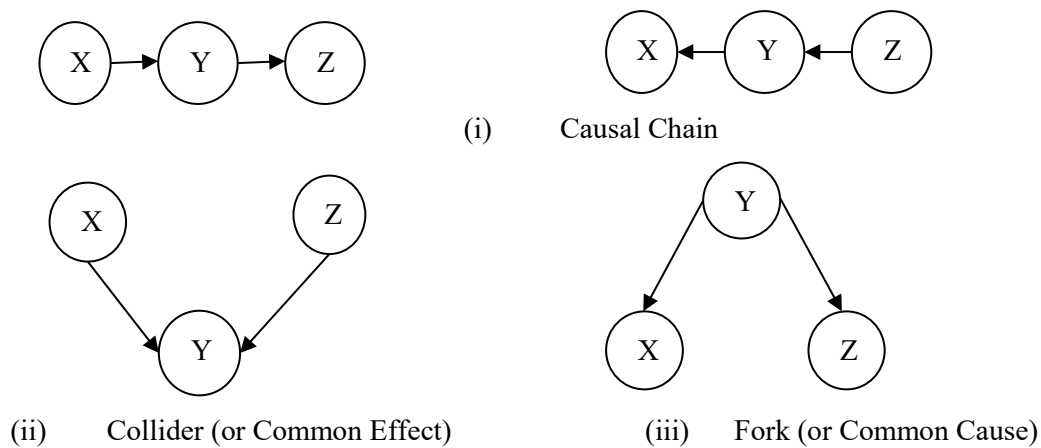

(i)  Causal Chain

(ii)  Collider (or Common Effect)   (iii)  Fork (or Common Cause)

**Fig. 3.** Representation of DAGs showing independence structure, $X \perp Z|Y$.

## 2.4 Counterfactual explanation

Potential outcome framework (Neyman, 1923; Rubin 1974) are expressed through the subscript notation for events and variables. For example, $Y_x(u)$, $Z_{xy}$ or Y (0), Y (1) or Z (x, y) are some of the notation used to express potential outcomes. The expression $Y_x(u)$ indicates that if u is randomly chosen, then $Y_x$ becomes a random variable, and it is represented as probability P ($Y_x$=y). Pearl (1995) used alternative expression for expressing $Y_x(u)$ which is of the form P (Y = y | set (X = x)) or P (Y = y | do (X = x)), which denotes the probability that event (Y=y) would occur if the X=x is provided as the treatment.

Let X and Y be two variables, X is called treatment and Y is called "response" or "outcome" and $M_x$ indicates modified version of reality M with X=x. Considering X as binary random variable, X=1 means that X is given treatment and X=0 is not provided any treatment. If $Y_1$ denotes the outcome if X=1 and $Y_0$ denotes the outcome if X=0.

$$Y = XY_1 + (1 - X)Y_0 \qquad (1)$$

In Eq. 1, random variables (X, Y) are replaced by (X, $Y_0$, $Y_1$, Y). If X=1 and only $Y_1$ is observed and $Y_0$ is not observed, but factually it must be $Y_0$ should have been observed. Here $Y_1$ is an unobserved variable leading to counterfactual. In context of counterfactual, the tuple of ($Y_0$, $Y_1$) is called potential outcome. Table 2 shows the representation of counterfactuals using the Eq. 1 and for 4 possible values of (X, Y), we get 8 different unobserved values, 4 each for $Y_i$, where i= {0,1}.

**Table 2.** Representation of counterfactuals for (X, $Y_0$, $Y_1$, Y),

* represents unobserved variables.

| X | Y | $Y_0$ | $Y_1$ |
|---|---|---|---|
| 0 | 0 | 0 | * |
| 0 | 1 | 1 | * |
| 1 | 0 | * | 1 |
| 1 | 1 | * | 1 |
| 0 | 1 | 1 | * |
| 0 | 1 | 1 | * |
| 1 | 1 | * | 1 |
| 1 | 1 | * | 1 |

In general, the definition of counterfactual $Y_x(u)$

$$Y_x(u) = Y_{M_x}(u) \qquad (2)$$

Eq. 2 enables answers to a large number of hypothetical questions which are of the form "What would Y be had X been set to x?" using scientific understanding of reality M. The same definition holds true for $M_x$ which is the modified model of reality, and all the elements of X are replaced by some constant. This substantially increases the number of counterfactual statements that may be computed by a given model and poses an intriguing question: How can a basic model, consisting of only a few equations, assign values to so many counterfactuals?

$$E(Y|do(X = 1), Z = 1) = E(Y|do(X = 0), Z = 1) \qquad (3)$$

$E(Y|do(X = 1), Z = 1)$ invoke post intervention events and it is expressible in the form of do(x) notation.

$$E(Y|do(X = 1), Z = z) = \frac{P(Y = y, Z = z|do(X = 1))}{P(Z = z|do(X = 1))} \qquad (4)$$

Counterfactuals (also known as "contrary-to-fact" representations) are what cognitive psychologists refer to as such mental images of potential futures for past events. Therefore, counterfactual thinking is the activity of "thinking about past possibilities and past or present impossibilities" (Roese, 1997). Counterfactual reasoning, on the other hand, is the process of creating an alternative scenario to the one that occurred and considering its ramifications (Pereira, & Machado, 2020). Additionally, it is asserted that a crucial mechanism for explaining adaptive behavior in a changing environment is counterfactual reasoning (Paik, et al.,.2014; Zhang et al., 2015). Counterfactuals are descriptions of things that did not happen or states of things that go against what is known to be true about the world (Kulakova, et al., 2013).

### 2.4.1. Counterfactual generation as an optimization problem

The first to suggest an optimization-based method for producing counterfactual explanation (CEs) is Wachter et al. (2017). The goal is to identify at least one CE $\tilde{x}$ that is the closest to the original factual instance x, so that $h(\tilde{x})$ equals a different target $\tilde{y}$, for a given trained classifier $h(.)$. The mathematical optimization model yields such a CE as given in Eq. 5.

$$\min_{x} \max_{\lambda} \lambda(h(x) - \tilde{y})^2 + d(x, \tilde{x}) \qquad (5)$$

where $\lambda$ is non-negative constant, d (.,.) is the Euclidean distance between original feature and generated feature.

According to Maragno (2022), who has compiled from different set of papers, it is necessary but not sufficient to take all seven criteria that should be fulfilled for generation of counterfactual explanations:

- **Proximity:** Regarding the feature values, the CE should be as near as feasible to the factual occurrence.
- **Coherence:** To get clear explanations, one-hot encoding for categorical data should allow us to map it back to the input feature space.
- **Sparsity**: The CE should be different from the factual case as sparsely as possible.
- **Actionability:** There are 3 different types of features for CE (i) immutable, (ii) changeable but not actionable, and (iii) actionable features.
- **Closeness of the data manifolds**: The produced CEs should closely resemble the observed (training) data to guarantee the production of plausible and useful explanations.
- **Causality:** Any (known) causal links in the data should be acknowledged in the suggested CEs to further provide plausible explanations.
- **Diversity:** Any technique for generating CEs should provide a collection of CEs that differ in at least one attribute.

## 2.5. Causal Effect Estimation

Back-door paths refer to undirected paths with an arrow into variable X that can introduce confounding. In a DAG G, a set of variables S is said to satisfy the back-door criterion with respect to an ordered pair of variables ($X_i$, $X_j$) if two conditions are met: (i) none of the nodes in S are descendants of $X_i$, and (ii) S blocks every path between $X_i$ and $X_j$ that includes an arrow pointing into $X_i$.

Mathematically, when S meets the back-door criterion is represented as:

$$P(Y|do(X = x)) = \sum_{s} P(Y|X = x, \ S = s) \, P(S = s) \qquad (6)$$

It is important to note that all the elements on the right-hand side of Eq. 6 are observational conditional probabilities, not counterfactuals. This implies that the criterion allows for identifiability and provides a strategy for adjustment.

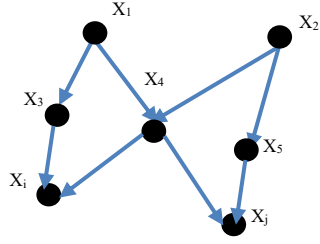

**Fig. 4.** Diagram representing Backdoor criterion.

The concept of back-door paths and the back-door criterion are illustrated in Fig. 4, where adjusting variables pairs $\{X_3, X_4\}$ or $\{X_4, X_5\}$ will yield estimate of Eq. (6). For a set of variables M to satisfy the front-door criterion, it must block directed paths from X to Y, have no unblocked back-door paths from X to M, and have X blocking all back-door paths from M to Y.

$$P(Y|do(X=x)) = \sum_{m} P(M=m|X=x) \sum_{x'} P(Y|X=x', M=m) P(X=x') \quad (7)$$

where X, Y and M are all observed variables. Variable M is the mediator between variables X and Y.

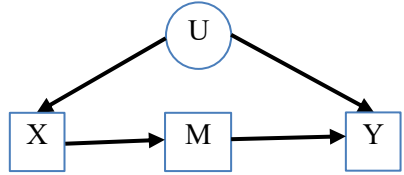

**Fig. 5.** Diagram of Front door criterion (Adapted from Pearl, 2009).

Fig. 5 represents the front door criterion, where U is the unobserved common cause of both X and Y. For front door criterion, apply back door criterion twice. In a causal relationship where X and Y share a common cause U ($X \leftarrow U \rightarrow Y$), a back-door path exists that introduces confounding, making it challenging to isolate the direct effect of X on Y. The effect of variable M on variable Y is influenced by the back-door path $M \leftarrow X \leftarrow U \rightarrow Y$. However, this path is blocked by variable X. Therefore, back-door adjustment can be employed to determine P (Y | do (M = m)), and we can directly find P (M | do (X = x)) = P (M | X = x). By combining these findings, we obtain P (Y | do (X = x)).

Instrument variable (IV) is used to identify the causal effect of X on Y. It satisfies conditions of independence from X and Y when controlled variables S are considered. This concept originates from econometrics and was initially employed to identify parameters in simultaneous equation models. By adjusting for the control variables, we can estimate the probabilities P (Y | do (I = i)) and P (X | do (I = i)) by using Eq. 8. The instrumental variable approach is useful in tracing the causal influence of I on Y through X (Pearl, 2009).

$$P(Y|do(I = i)) = \sum_{x} P(Y|do(X = x)) P(X = x \mid do(I = i)) \qquad (8)$$

where variable I represent the Instrument variable.

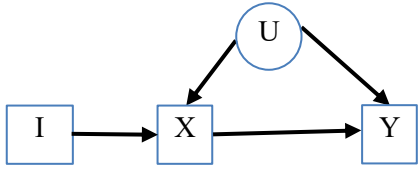

**Fig. 6.** Diagram of Instrument variable.

In the instrumental variable setup depicted in Fig. 6, where I act as an instrument for estimating the effect involving variables X, Y, the identification of the causal effect of X on Y, under the assumption of linearity, is equivalent to determining the coefficient associated with the X → Y arrow.

### 2.5.1. Estimation of Average Causal Effects

When we determine the effect of variable X on variable Y using the back-door criterion with control variables S, Eq. 9 represents the expected value of Y given that we intervene on X by setting it to the value x, and it can be computed by summing the product of each possible outcome y with the corresponding probability of Y being equal to y when X is set to x.

$$\mathbb{E}(Y|do(X = x)) = \sum_{y} y P(Y = y|do(X = x)) \qquad (9)$$

Eq.9 can be rewritten by substituting Eq.6 in P (Y=y | do (X=x)) as follows:

$$\mathbb{E}(Y|do(X = x)) = \sum_{y} y \sum_{s} P(Y|X = x, \ S = s) P(S = s) \qquad (10)$$

Some of the terminologies of treatment effect are described as follows:

1. **Individual Treatment Effect (ITE):** Individual Treatment Effect (ITE) refers to the causal effect of a specific treatment or intervention on an individual unit or subject. It quantifies the difference in outcomes that can be attributed to the treatment compared to the counterfactual scenario where the individual did not receive the treatment. The ITE is denoted as $\tau_i$ and can be formulated as $\tau_i = Y_i(1) - Y_i(0)$, where $Y_i(1)$ represents the potential outcome when the individual receives the treatment, and $Y_i(0)$ represents the potential outcome when the individual does not receive the treatment.

**2. Average Treatment Effect (ATE):** Average Treatment Effect (ATE) represents the average causal effect of a treatment or intervention on a population. It measures the difference in outcomes between the treated group and the control group. The ATE is denoted as $\tau$ and can be calculated as the average of the individual treatment effects: $\tau = E[\tau_i]$, where $E[.]$ denotes the expectation over the population.

**3. Conditional Average Treatment Effect (CATE):** Conditional Average Treatment Effect (CATE) captures the heterogeneous treatment effects across different subgroups or conditions. It quantifies the causal effect of a treatment for a specific subgroup, given the individual's characteristics or covariates. The CATE is denoted as $\tau(x)$ and can be expressed as $\tau(x) = E[Y_i(1) - Y_i(0) | X = x]$, where X represents the individual's covariates or characteristics.

**4. Local Average Treatment Effect (LATE):** Local Average Treatment Effect (LATE) (Imbens & Angrist, 1994) is a causal effect that specifically applies to individuals who are compliers in a treatment assignment. Compliers are individuals whose treatment status changes based on external factors, such as an instrument variable. The LATE estimates the causal effect of treatment for compliers only. It is denoted as $\tau_c$ and can be calculated as
$\tau_c = E[Y_i(1) - Y_i(0) | Z_i = 1]$, where $Z_i$ represents the instrument variable indicating compliance.

# 3. Statistical Underpinnings of CI

In this section, we will explore Bayesian Causal Network (BCNs) and Granger causality. BCNs can be address variety of causal inference questions, such as: What is the effect of changing one variable on another variable? What are the direct and indirect effects of one variable on another variable? Granger causality can be used to identify potential causal relationships between two time series, but it cannot be used to determine the direction of causality or to identify indirect effects.

## 3.1. Bayesian Causal Network

To aid in reasoning under uncertainty, Pearl developed a type of probabilistic graphical model known as a Bayesian Network (BN) (Pearl, 1995; Neufeld, 1993). Probability theory offers a way to express and control the relevance, causality, conditionality, and likelihood relationships that are fundamental to human reasoning (Neufeld, 1993). To answer questions like "*what will be the likely consequence of this intervention?*" or "*what components are related to this effect?*" Bayesian Networks combine probability theory with graph theory, causation, and other concepts.

BN is a tuple of Directed Acyclic Graph (DAG) G and set of parameters to represent the strength and shape of relationships between variables denoted by Θ (Kitson et al., 2021). The BN is a special DAG that helps us to think about intervention if we assume this causal relationship. A directed edge between two nodes in a graph indicates that node B is a child of node A, or vice versa.

The joint probability distribution over all the variables, P(X), is represented by the BN. It contains a list of conditional dependencies as well as indirect conditional independence. Following are the 2 assumptions made about the DAG used in BN:

**(i) Markov condition:** Given its parents, each variable X in G is conditionally independent of all its non-descendants. Owing to this condition, the joint probability distribution P(X) can be broken down as given in Eq. 6 (where Pa (Xi) are Xi's parents).

$$P(X_i) = \prod_{i=1}^{n} X_i \tag{11}$$

**(ii) Minimality condition**: Since P(X) does not contain a conditional independence; those edges cannot be removed from the DAG without the graph indicating one.

## 3.2. Granger causality

The notion of causality, developed by Granger (1969) is acknowledged largely in the literature on econometrics. Granger causality does not entail true causality. The main principle of the idea is that causes come before effects and can aid in forecasting them. It is also presumable that the cause contains information about the consequence that cannot be obtained in any other way.

Let $H_{t-1}$ be the history of all relevant information till time t-1. P ($x_t$ | $H_{t-1}$) is the probability for prediction of $x_t$ given $H_{t-1}$. According to Granger, it is stated that y can be causal for given x for the given condition.

$$\sigma^2(x_t - P(x_t \mid H_{t-1})) < \sigma^2(x_t - P(x_t \mid H_{t-1} \setminus y_{t-1})) \tag{12}$$

Eq. 12 indicates that by including the history of y, the variance of the optimal prediction error of x is reduced. We can also say that y is "causal" of x if past values of y improve the prediction of x. This portrayal is obviously founded on consistency and doesn't (straightforwardly) highlight a causal impact of y on x: y working on the expectation of x doesn't mean y causes x. In any case, accepting causal impacts is requested in time (i.e., cause before impact), that's what granger contended, under certain suppositions, if y can foresee x, then, at that point, there should be a causal effect (Shojaie & Fox, 2021).

There was no mention of the potential instantaneous connection between $x_t$ and $y_t$ in the concept of Granger causality. We refer to instantaneous causation as existing when the innovation to $y_t$ and the innovation to $x_t$ are connected. Instantaneous correlation between two time series is typically (or at least frequently) found, but since the causality (in the "*actual*" sense) can be either positive or negative, one typically does not test for instantaneous correlation.

The identification of a singular linear model was the foundation of Granger's initial argument. indicating the time-varying variable vector t by $x_t = (x_{1t}, x_{2t}, \cdots, x_{pt})^T$. He considered following the linear model as expressed by Eq. 12, in which he stated that this model is generally not identifiable unless the matrix $A^0$ is diagonal is given by Eq. 13.

$$A^0 x_t = \sum_{i=1}^{d} A^k x_{t-k} + e_t \tag{13}$$

where $A^0$, $A^1$,.., $A^d$ are p x p matrices comprising coefficients, d is the lag (or order) in time series data, error term $e_t$ is of p-dimensional, it can be either diagonal or non-diagonal covariance matrix $\sum$.

Overall, Granger causality is a useful tool for identifying potential causal relationships between time series data. However, it is important to remember that Granger causality does not provide definitive proof of causality. For more complex causal queries, causal inference methods that go beyond Granger causality are needed.

## 4. Extended CRISP-DM by incorporating Causal Inference

The use of causal inference in the CRISP-DM process can help ensure that the data mining project produces results that are relevant to the business problem and that are not simply spurious correlations. Here is an example of how CRISP-DM and causal inference could be used in the BFSI industry:

**Business Understanding:** The bank would first need to understand the business problem they are trying to solve. In this case, they might want to reduce the number of loan defaults.

**Data Understanding:** The bank would then need to collect data on all their customers, including their loan history, financial information, and demographic data.

**Data Preparation:** The bank would need to clean and prepare the data for analysis. This might involve removing duplicate records, imputing missing values.

**Modelling:** The bank could select the potential machine learning algorithms and build the model on the prepared dataset. Once the model is generated, it will be evaluated on the chosen metrics according to the problem statement.

**Evaluation:** The bank would then need to evaluate the results of their models to make sure that they are accurate and reliable. They might do this by comparing the results of their models to historical data or by conducting a pilot study.

**Deployment:** Once the bank is confident in the results of their models, they could deploy them to production. This would allow them to use the models to make decisions about which customers are most likely to default and to develop targeted interventions to prevent defaults.

**XAI and CI:** The bank would then use causal inference methods to estimate the causal effect of different factors on loan defaults. For example, they might estimate the effect of credit score, income, and debt-to-income ratio on the likelihood of default.

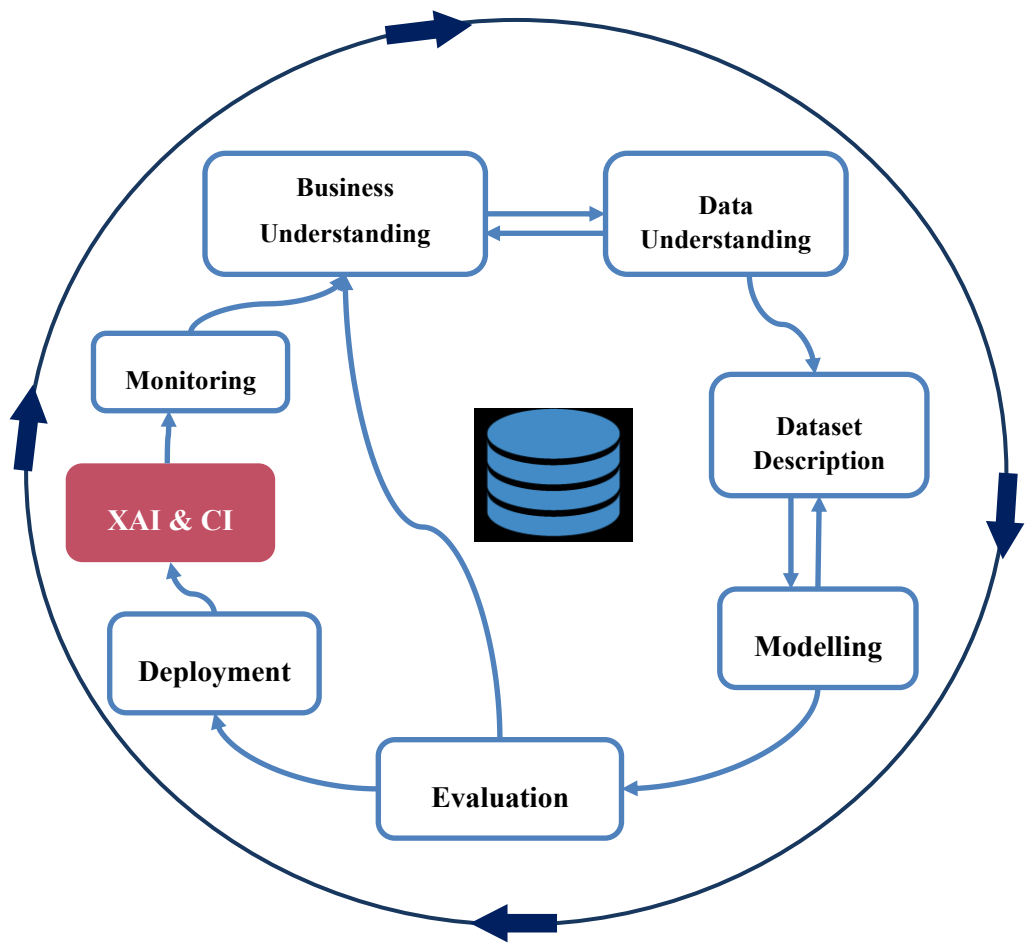

**Fig. 7.** Extended CRISP-DM methodology with inclusion of interpretability and causality

It is worth noting that our idea of extending CRSIP-DM is different from that of Plumed et al., (2021) in that we recommend to invoke causal inference only after deployment so that its value addition can be clearly seen by the end users.

Hernan et al., (2019) discuss counterfactual prediction as a new type of data science problem that sheds light on how causal inference from data might be addressed using data science. The CRISP-DM process works well for this task and others that execute causal inference (under the modeling step), although expert knowledge and experience become critical (and as a result, the inner stages of the CRISP-DM process are more difficult to automate). The expert's causal knowledge in the form of graphical models, together with other types of domain knowledge or extracted patterns, may be used to generate new data, such as through randomized trials or simulations on the observed or generated data (Plumed et al.,2021).

Fig. 7 depicts the extended CRISP-ML/DM methodology including interpretability and causality. It involves two stages. In stage 1, the team assesses the interpretability and causality requirements, which allows them to make an informed decision on which ML modeling and interpreting techniques to use. This decision should increase the likelihood of achieving project objectives and meeting stakeholder expectations in terms of accuracy, interpretability, and causality. In stage 2, the team designs and develops the chosen ML model, with a focus on interpretability and causality to facilitate explanation and understanding of the model's output (Kolyshkina, & Simoff, 2021).

Knowing fully well that the CRISP-DM, CRISP-ML are independent of the data volume, variety, and velocity, we propose this extended methodology CRISP-DS for data science, where data science turns out to be a superset of DM which in turn includes ML, statistics, operations research and databases (AI in Banking: A Primer, 2020)

## 5. Review of papers

Fig. 8 shows the distribution of papers published in BFSI from the year 1992 to year 2023 that use different causal inference methods. The literature survey is grouped on 2 themes, first theme is on selecting papers related to domains making use of causal inference in Banking, Finance, and Insurance. Finance domain is further categorized into (i) corporate finance, which include articles which includes the following keywords as investments, assets in abstract or in title, (ii) financial economics, (iii) financial governance include keywords as financial risk, financial policy, papers describing granger causality for time series,(iv) Behavioral finance (v) accounting, (vi) Time series, and (vii) application of financial services such as churn modeling and credit scoring, credit lending. Another category for this survey paper is based on the distribution of papers on the causal inference methods used either implicitly or explicitly that includes statistical methods such as (i) Bayesian Network and (ii) Granger Causality, (iii) counterfactuals, and (iv) explainability for credit scoring, churn modeling. As these survey papers had many overlapping distributions of papers and to resolve such conflicts it has grouped into that category that has resulted in maximum word counts from that paper. For e.g., suppose a paper has counterfactuals and corporate finance, and the paper is more inclined towards counterfactual than corporate finance, that paper will be included in counterfactual category rather than corporate finance category. It shows that in the last 5 years more papers have been published in BFSI.

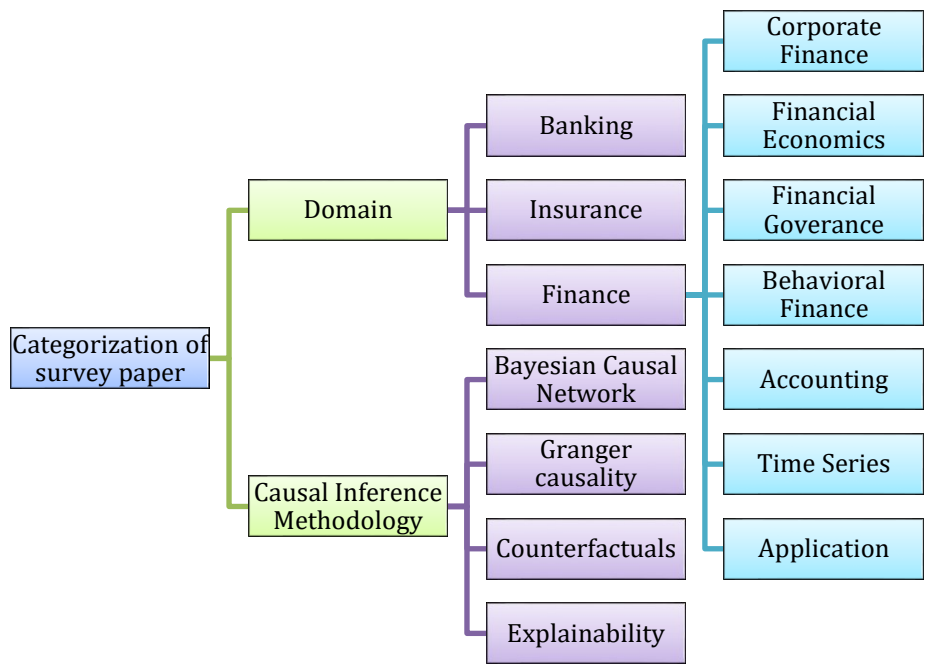

**Fig. 8.** Categorization of survey papers into two different themes.

Table 3 presents the distribution of journal articles. Table 4 shows the distribution of published articles under different conferences along with their abbreviations and number of papers published in each of them. Fig. 9 shows the distribution of papers published in BFSI from the year 1992 to year 2023 that use different causal inference methods. From this Fig., one can clearly see that in the last 5 years, more papers have been published in BFSI.

The remaining 9 articles are published book chapters and in preprint such as ArXiv and Social Science Research Network (SSRN) which is as presented in Table 5. Out of which 2 are published in different book chapters and 4 are available in ArXiv and 3 are available in SSRN.

**Table 3.** Distribution of published articles in Journals.

| Journal Name | No. of papers |
| --- | --- |
| Banks and Bank system | 1 |
| Critical Finance Review | 1 |
| Accounting Research | 1 |
| Historical Social Research | 1 |
| Operational Research Society | 1 |
| Financial Markets and Portfolio Management | 1 |
| BIS Working Paper | 1 |
| Oxford Bulletin of Economics and Statistics | 1 |
| Physica A: Statistical Mechanics and its Applications | 1 |
| Journal of Forecasting | 1 |
| Economics and Finance | 1 |
| Acta Psychologica | 1 |
| Network Theory in Finance | 1 |
| Applied Informatics | 1 |
| Philos Trans A Math Phys Eng Science | 1 |
| Accounting, Organizations and Society | 1 |
| International Review of Economics & Finance | 1 |
| Microeconomics: Intertemporal Firm Choice and Growth | 1 |
| Knowledge and Information Systems | 1 |
| Review of Financial Studies | 1 |
| **Total** | **20** |

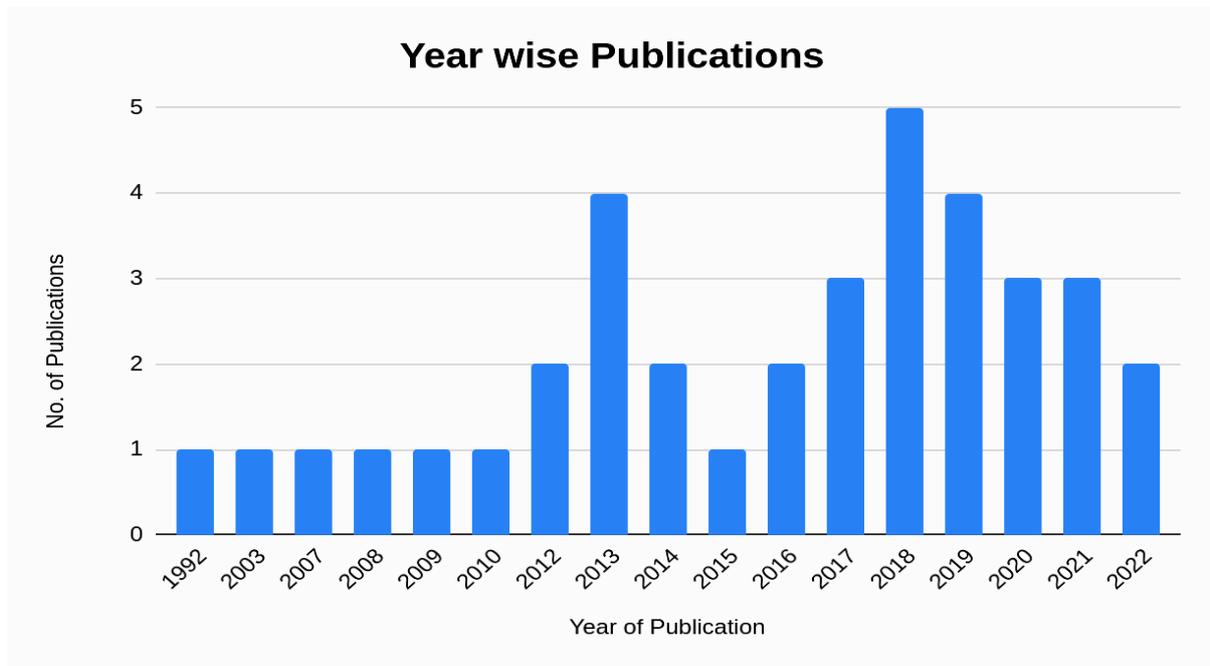

**Fig. 9.** Year wise (1992-2023) distribution of articles published.

.

Table 4. Distribution of published papers in Conferences.

| Conference Name | Abbreviations | No. of papers |
|---|---|---|
| International Conference on Machine Learning and Applications | ICMLA | 1 |
| International Conference on Big Data | Big Data | 1 |
| Conference on Computational Linguistics | COLING | 1 |
| Conference on Neural Information Processing system | NeurIPS | 1 |
| International Conference on Computational Science | ICCS | 1 |
| International Joint conference on Artificial Intelligence | IJCAI | 2 |
| International conference on Machine Learning | ICML | 1 |
| Extending Database Technology | EDBT | 1 |
| **Total** | | **9** |

**Table 5.** Distribution of published papers in Preprint and Book Chapters.

| Type of Publications | Title of publication | Total |
|---|---|---|
| Book Chapter | Applied Bayesian Modeling and Causal Inference | 1 |
| | Bayesian Methods in Finance | 1 |
| Preprint | ArXiv | 4 |
| | SSRN | 3 |

Fig. 10 illustrates the different articles which have been published under the distribution of different publishers. 25% of paper published are through Elsevier, followed by Springer (9%) and IEEE (9%) which includes several papers from journals as well as from conferences.

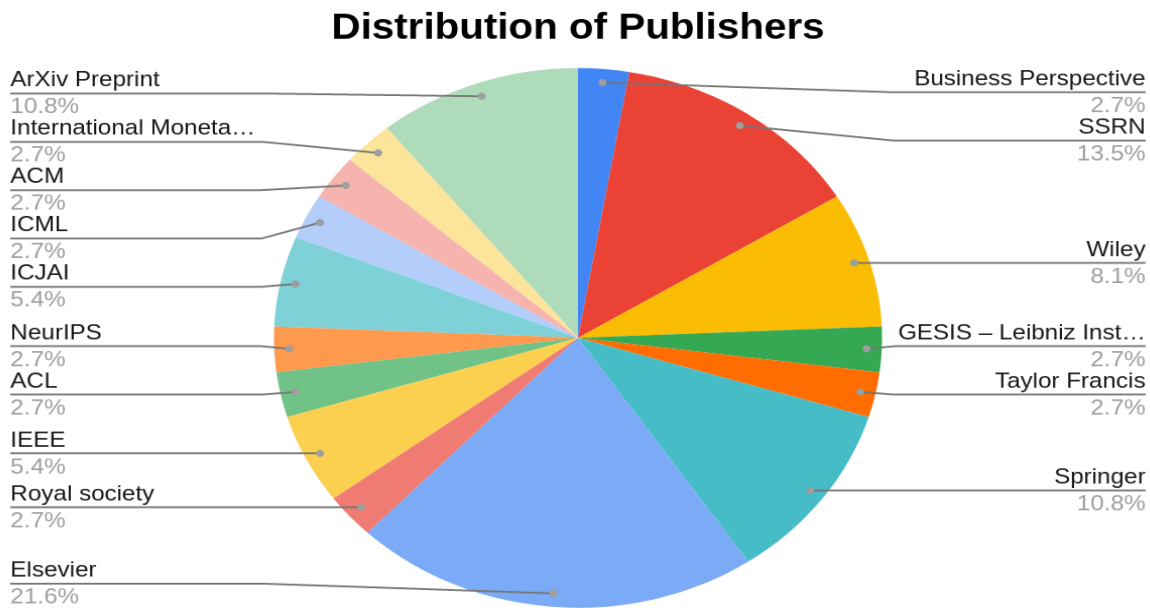

**Fig. 10.** Distribution of articles published by various publishers.

## 5.1. Survey based on domains

In this section, we survey the papers, which are collected following the filtration scheme depicted in Fig. 1). We divided the research papers into different groups based on their domain.

### 5.1.1. Banking

Michail (2019) examined the effect of negative interest rates on inflation and bank lending in three European countries namely Denmark, Sweden, and Switzerland. He concluded that negative interest rates would not prevent banks from lending freely. The effect of interest was obtained using a simple regression, which includes the regressor of interest (i.e., the policy variable) and excludes other control variables. This study was carried out with the following assumptions: (i) The counterfactual scenario for the interest rate should remain at exactly zero during the whole post intervention period. (ii) policy rate remaining at zero for a long time can weaken its homogeneity to the economy. It leads to the conclusion that the policy's ineffectiveness lies within the indisputable fact that negative interest rates did not ease the case for the factors limiting the provision of bank lending; particularly bank funding prices, and return-on-Equity.

Kolodiziev et al., (2014) developed an integrated approach for monitoring the banking system stability using causal analysis. This was carried out for Ukrainian banks with four groups of indicators assessing the intensity of credit and financial interaction in the interbank market, the efficacy of the banking system functions, structural changes and financial disproportions in the banking system, and activities of systemically important institutions. The hypotheses were formulated for the presence of causal relationships between groups of indicators. They built models of linear and polynomial regression that can show the dependency of the values of the banking system stability index. When causal analysis is used to monitor the stability of the financial system, the most important components can be determined by organizing the causal linkages between the indicators for its evaluation. This indicator illustrates both individual indicator deviations and changes in the subindices, demonstrating the interdependence of the stability of the financial system.

Stolbov & Shchepeleva, (2020) examined the causal relationships between systemic risk, economic policy uncertainty, and firm bankruptcies. The analysis is conducted by considering global volatility measured by the VIX index as a conditional factor. The study employed various methods to test for Granger causality in both the time and frequency domains. The authors argue that the impact of systemic risk, economic policy uncertainty, or the VIX index on bankruptcies is contingent upon the magnitude of deleveraging by banks in relation to the private nonfinancial sector.

### 5.1.2. Finance

Finance has many different subsets of domains such as corporate finance which deals with capital investment of the companies, financial economics which lies at the intersection of finance and economics, finance governance that deals with making policies and management of those policies, operations, accounting, and time series.

### 5.1.2.1 Corporate Finance

Gong et al. (2019) analyzed the network topology using the centrality indicator. From the perspective of time and space dimensions, the causal complex network among financial institutions is analyzed with the dynamic variation of systemic risk measured. Closeness and eigenvector centrality were the next steps in using the centrality techniques. Three different systemic risk metrics, namely systemic risk index (SRISK), the marginal expected shortfall (MES) and the conditional Value at Risk (CoVaR) were employed. The banks, securities, and insurance businesses in China's financial markets are using the recommended methodology framework. Financial entities serve as the vertices and causality orientations as the edges of the causal network. When systemic risk increases, the suggested network framework can offer a reliable instrument for early warning.

### 5.1.2.2 Financial Economics

Davis (2023) showed that there are twelve models found in literature that show significantly inelastic demand, unlike classical models. This is because it is challenging to trade against price changes in real-world situations, which confirms the estimates of demand elasticity. There are two types of predictors, (i) direct function which is dependent on price ratio variable, and (ii) independent of price. Counterfactual experiments conducted thereof derived 62 anomalies only for incumbent investments, where rank correlation rather than standard correlation was employed.

### 5.1.2.3 Financial Governance

Using causal graphs, graph visualization, and graph anomaly detection, Ravivanpong et al. (2022) determined the underlying causes of a change in risk profile or limit breach. A risk manager at an asset management firm is required to keep track of 100 to 250 portfolios each day. Over 160 risk factors are combined into each of these portfolios. A causal relationship between underlying assets that is missed by conventional approaches or ignored by risk and portfolio managers can unexpectedly result in a significant correlation between many portfolios of various sorts during a minor crisis. Agglomerative hierarchical clustering (AHC) of portfolio risk profiles with network visualization is used as a simpler and practical alternative. Effective Transfer Entropy (ETE) is used for the non-parametric method that can detect both linear and non-linear statistical dependencies. The combination of AHC analysis and Value at Risk (VaR) exposure rates is replaced using causal graphs in terms of practicality and technical requirements.

### 5.1.2.4 Behavioral Finance

Rigana et al., (2021) devised a new measure for contagion between individual currencies within the foreign exchange market and shows how contagion paths within Forex work using causal inference. Network contamination is based on causal graph model theory. The N assets that are to be analyzed and their log returns are represented as multivariate random variables.

Tsapeli et al., (2017) demonstrated the causal impact of social media sentiment on traded assets. Stock market returns are influenced by emotional, social and psychological factors. Pure correlation analysis is not enough to prove that stock returns are affected by such emotional factors. Causal impact of information posted in social media on stock market returns of 4 big companies namely Apple, Microsoft, Amazon, and Yahoo was studied. Tweets are collected using the names and stock symbols. Its result is compared with a multivariate Granger causality model and with an information theoretic approach based on Runge's framework. It has been found out that Twitter data polarity does indeed have a causal impact on the stock market prices. The disadvantage of this approach is based on observational data rather than experimental procedures that could result in bias in case of missing confounding variables. Performing similar analysis in the B2B sector is an open question, which can be explored.

### 5.1.2.5 Accounting

Event research is one of the most used methods in accounting and financial studies. Castro (2017) developed a framework that can relate estimation methods in finance with the contemporary approaches to causal events. The time leading up to the event and the outcome variable of interest, typically stock holding period returns, are the standard components of conventional finance event studies. It is customary to define the event window as an interval around the event $[T_{0-d}, T_{0+d}]$, where d is the number of days surrounding the event.

### 5.1.2.6 Time Series

Analyzing the causal impact of rare events is highly essential because the impact caused by them is the most crucial one. This is very critical in finance and other fields such as bioinformatics, and computational sciences. Hence, Kleinberg (2013), proposed an assessment of rare causes (ARC) method to infer the causation of the rare event which improves decision-making. ARC first calculates the unexpected and conditional unexplained values and infers the normal model. Later, an unpaired t-test is performed to determine the statistical significance of each pair of relationships. ARC is tested on two different scenarios, where (i) the rare event influence is constant, and (ii) the rare event influence is functional in nature.

Kleinberg et al. (2010), devised a new algorithmic framework for inferring the causal relationship and applied this framework for time series data. In this proposed framework, causal relationships are represented by logical formulas for testing arbitrarily complex hypotheses in a computationally efficient way. A causal relationship in a propositional probabilistic branching time temporal logic (PCTL) accumulates all the price data at varying time scales from interest rate announcements to earnings reports to news stories and even tweets.

Moraffah et al., (2021) focused on two causal inference tasks, i.e., treatment effect estimation and causal discovery for time series data. The challenges of causal treatment impact estimation, based on the current approaches, are on various situations at the time, and describe the cutting-edge approaches in each category. Time-invariant treatment effect is defined as the treatment occurring at a single moment in time and remaining constant after that, it is said to be time-invariant or fixed, whereas time-varying treatment effect is used in Granger causality and conditional independence-based approaches.

Peters et al., (2013) studied a class of restricted structural equation models for time series known as the Time Series Models with Independent Noise (TiMINo). In contrast to conventional approaches like Granger causality, TiMINo requires independent residual time series. Data from time series are used as the input, and the output is either a DAG that calculates the summary time graph or is ambiguous. The authors modified the approach for additive noise models, for time series without feedback loops. It can be applied to model unconditional independence that can be used with multivariate, linear, nonlinear, and instantaneous interaction data. Additionally, TiMINo considers identifiability that results from constrained structural equation models.

Chikahara & Fujino, (2018) employed supervised learning framework classifiers rather than regression models for time series causal inference. The distance between the conditional distributions is determined using a feature representation and historical variable values. Multivariate time series can be added to the classifier framework. Following datasets were used in for bivariate time series (i) Synthetic dataset: 15, 000 pairs of synthetic time series with the length T = 42 (ii) Real world dataset Five pairs of bi-variate time series includes cause-Effect Pairs database and for Multivariate time series:

Geiger et al., (2013) has presented 2 estimating techniques and assessed those using simulated and real-world data. The two algorithms are specifically designed to address conditions 1) non-Gaussian, independent noise or (2) no influence from X to Z, respectively. A first order vector autoregressive (VAR) process is formed by X and some hidden unmeasured time series Z, with transition matrix A. Firstly, it was investigated that the prerequisites for identifying the underlying system's causal features from the provided data. Later, estimating techniques were suggested, and it was demonstrated that they operated on simulated data under the corresponding conditions.

**5.1.2.7 Miscellaneous Applications**

Rudd et al., (2017) proposed a churn prediction system, which can be used for almost all non-subscription business settings. In this framework, first, the churn prediction score is calculated by using multi-layer perceptron (MLP) followed by predicting the possible causes by performing a causal analysis. To perform causal analysis structural equation models (SEMs) and Counterfactual based models were employed to identify the reasons for the customer to churn out. Along with the above, they proposed a novel feature engineering process, which is based on recency, frequency and monetary (RFM) aspects of the customer interaction that is used to identify important customers. By employing this novel feature engineering technique, significantly improved the performance of the model.

The decision-making in credit scores suffers majorly from the following two (i) selection bias and (ii) limited historic testing. This motivated Fahner (2012) to propose an approach, which can handle multiple ordinal or categorical treatment effects. Using this approach, following advantages were observed (i) extracts the local information by utilizing the support regions, (ii) the global model is derived by using the local information. The authors tested their proposed framework on risk-based pricing and credit line increase problems. In both datasets, the complex causations were obtained by using the proposed model.

**5.1.3. Insurance**

Guelman & Guillén, (2022) applied the causal effect of a rate change on the policyholder's lapse. It is assumed the inclusion of policyholder's lapse which covariates adjusts with the potential exposure of correlations between price elasticity and other explanatory variables. The model is built with a series of lapse probability models, one for each rate change level trained on GBM which has a built-in variable selection procedure. For causal effect estimation, propensity scores and matching algorithms were used for finding pairs of policyholders exposed to distinct rate change levels. Lastly, counterfactual outcomes obtained from the matched pair are inferred. This model would assist managers in selecting an optimal rate change level for each policyholder for the purpose of maximizing the overall profits for the company. Table 5 displays the distribution of articles based on domains grouped according to number of journal articles, number of conferences, preprint.

## 5.2. Survey based on causal inference methodologies

In this section, we divided the research papers based on causal inference methodology namely Statistical-based methods such as Bayesian Causal Network and Granger causality, Application of causal inference through counterfactuals and. Explainabillity-in financial services that includes churn prediction, Credit card scoring etc.

### 5.2.1. Bayesian Causal Network

Jacquer & Polson, (2012) surveyed about the various Bayesian econometric methods in finance. It focuses on the application related to Markov Chain Monte Carlo (MCMC) and particle filtering (PF) algorithms. MCMC allows the study of interference for complex models and latent variables by considering the posterior probabilities. MCMC is discussed in the view of how it handled the stochastic volatility (SV). PF algorithm helps in comparing the various methods which are discrete over time. Bayesian methods was applied to various applications optimal portfolio design, returns predictability, asset pricing, option pricing etc.

To overcome the heterogeneity and less availability of data related to operational risk modeling, Sanford and Moosa, (2012) developed a Bayesian network structure which is suitable for operational risk modeling. It was modeled as structured finance operating (SFO) units in an Australian bank. The authors incorporated the methodology which is performed in three different stages. (i) Structural development and evaluation. (ii) Probability elicitation and parameter estimation. (iii) Model validation. In the first step, all the relevant causal relations were identified without any elicited or learned parameters. All the dependencies among the nodes will be studied. The inputs will be taken from various domain experts who are not involved in the network design to identify comprehensive, unambiguous causal relations. In the second stage, marginal and conditional probabilities were calculated by reference to domain-based experiment results. Later, in the end, this constructed model is evaluated by conducting various methods such as (i) an elicitation review, (ii) sensitivity analysis and (iii) case evaluation.

Gao et al., (2017) proposed a novel stress testing framework by combining Suppes-Bayes Casual Networks (SBCNs) and various classification algorithms. SBCN is a variation of the traditional Bayesian Causal Network and is of the probabilistic graphical model. SBCN is different from the traditional Bayesian Causal Network in the following way: (i) Probability causation am used to exploit better solutions. (ii) Prima facie casual relations among different variables are also captured along with the conditional independence. (iii) By using maximum likelihood estimation (MLE), all the spurious causes were removed. The authors conducted various experiments on various stress scenarios. In all the scenarios, the SBCNs had shown efficiency, in terms of computation and data usage.

### 5.2.2. Granger causality

Stavroglou et al., (2017) carried out the study in financial assets. Categorically, the researchers divided the methods handling financial assets into 8 different types. Among them linear inter temporal cross correlation (LICC) and nonlinear inter temporal cross correlation (NICC) are quite popular and largely applied. The authors urged the importance of studying the causal inferences and interventions which help to avoid financial debacles like financial crisis occurred during 2007-2009. The following financial assets are collected (i) stock market indices from major economic countries like USA, Japan, China, India etc., (ii) Government bonds from USA, China, Italy etc., (iii) Oil prices. The authors analyzed the time-tested causal relationships by using both the LICC and NICC. It turned out 50% of the links are common in both LICC and NICC and each method occupied its own importance. The authors showed the effect of increase in oil prices over the China stock market crash. The extant approaches mostly focus on machine learning prediction, but studying causation and intervention gives an invaluable asset in crucial times like financial crises. Hence, (Tiffin, 2019) conducted an empirical economic study on the impact of the financial crisis on growth. It was focused on casual random forest algorithms. Using this algorithm, the authors studied the risk pertaining to the variables and potential thresholds that need to be followed is observed, non-linearities and studied the potential role of exchange rate and how it plays a key-role in impacting the development of a country.

Eichler, (2013) discusses problems with spurious causality and approaches to tackle these problems. The author has discussed the difficulties with applying Granger and Sims causality empirically. Additionally, a learning identification approach that uses latent variables to learn causal time-series structures is used. Analyzing the causal impact of rare events because the impact caused by them is the most crucial one. This is very critical in finance and other various fields such as bioinformatics, and computational sciences.

### 5.2.3. Counterfactuals

Lundberg & Frost, (1992) for the first time studied the influence of counterfactuals in the marketing domain where the situation is always changing and unpredictable in nature. The authors conducted an empirical analysis and tested the hypothesis derived from Norm Theory in the context of trading. The authors describe that counterfactuals are helpful in making dynamic decision-making environments.

The authors discussed the importance of post-decision processing to analyze the effects caused by the previously constructed marketing strategy.

Svetlova, (2009) mainly pointed out that "counterfactual analysis is very important and should be treated as an element of social life and not only as the human mentality". The counterfactuals in the context of portfolio management and how it adversely affects the financial markets are studied here. Portfolio management is largely influenced by the following factors: fundamental factors, political factors, macro-economic factors etc. The authors mentioned that counterfactuals will help us to maintain the portfolios.

Brodersen et al. (2015) suggests a novel approach to estimate causal impact by utilizing a diffusion-regression state-space model. Unlike traditional difference-in-differences methods, state-space models offer several advantages. Firstly, they enable the inference of the temporal progression of attributable impact. Secondly, these models allow for the incorporation of empirical priors on the parameters using a fully Bayesian treatment. Lastly, they offer flexibility in accommodating multiple sources of variation, including local trends, seasonality, and the varying influence of contemporaneous covariates.

Gan et al. (2021) designed a model agnostic framework to generate viable counterfactuals for model risk management. The authors automated the workflow with the cloud native algorithms. They achieved the containerization and workflow orchestration by using kubeflow. The authors evaluated the performance of the designed tool on Freddie Mac dataset.

Wang et al. (2023) proposed new sparsity algorithm for finding counterfactual explanation as an optimization problem. By this algorithm sparsity is maximized for higher dimensional input to solve corporate credit rating of a company. Credit rating speeds up purchasing or issuing bonds. For classification of high dimensional inputs, function is not injective which was the major challenge faced in counterfactual explanations. By introducing a new sparsity algorithm, they minimized the number of features modified by small changes which will result in counterfactual explanations for those features.

### 5.2.4. Explainability approach

Corporate mergers and acquisitions (M&A), which represent annual investments in billions of dollars, present an intriguing and difficult field for artificial intelligence. Unfortunately, the new research on explainable AI (XAI) in financial text categorization has gotten little to no attention, and many of the algorithms now used to generate textual-based explanations produce very improbable explanations that undermine user confidence in the system. In order to solve these problems, Yang et al., (2020) proposed a novel methodology for creating plausible counterfactual explanations. A transformer variation must first be refined on the M&A prediction job, along with adversarial training. Following the prediction, key words in the test cases are determined using a sampling contextual decomposition technique. Third, by swapping out these terms with grammatically appropriate alternatives, a counterfactual explanation is produced. Extensive quantitative studies carried by authors showed that this technique not only increases the model's accuracy when compared to the state-of-the-art and human performance, but also produces counterfactual explanations that are noticeably more believable based on human trials.

Credit applications are predicted using commercial, interchangeable black-box classifiers, and individual forecasts are explained using counterfactual justifications. To favor counterfactuals acting on highly discriminative features, Grath et al., (2018) added a weight vector to the distance measure. To create these weight vectors, two ways are suggested. The first one is based on the analysis of variance (ANOVA) F-values between each characteristic and the target to determine the relevance of each feature globally. The second one uses a Nearest Neighbors technique to aggregate the neighborhood's relative changes in relation to x.

Dastile et al. (2022) employed a customized genetic algorithm to generate concise and sparse counterfactual explanations for predictions made by black-box models. They evaluated the effectiveness of their approach using publicly accessible credit scoring datasets. The explanation provided by their approach can also be applied to elucidate the reasoning behind approved loan applications. Additionally, enhancing the performance of the method can be achieved by developing an optimal fitness function through genetic programming that effectively captures various characteristics of counterfactual explanations. Furthermore, instead of focusing solely on explaining individual instances, elucidating the underlying workings of the black-box models can significantly enhance the transparency and explainability of credit scoring models.

Bueff et al., (2022) proposed a novel approach to address the lack of interpretability in machine learning models used for credit scoring by introducing the concept of counterfactual explanations, which provide insights into how changes in input variables can affect credit scoring outcomes. By generating counterfactual scenarios, the authors aim to provide a better understanding of the relationship between input variables and credit risk, thereby enhancing the interpretability of the models. The study utilizes a dataset of credit applications and employs various machine learning algorithms to develop credit scoring models. These scenarios simulate potential changes in the input variables, such as income or outstanding debts, and examine their impact on credit scores.

Table 6 provides the summary of survey papers by combining both the causal inference approach and domain and subdomain categories together for all papers. A new column describing the datasets used by authors in their respective papers is added.

**Table 6.** Summary of the surveyed papers

| References | Causal Inference approach | Input data type | Dataset | Domain# | Sub domain |
|---|---|---|---|---|---|
| Brodersen et al. (2015) | Counterfactual | Time series | Google's advertisers click rate in the United States | Finance | Time series |
| Bouezmarni et al., (2019) | Granger causality index | Time Series | | Finance | Time series |
| Castro, (2017) | Counterfactual | Time series | Merger & Acquisition (M&A) data from the Thomson Reuters SDC Platinum Financial Securities database | Finance | Accounting |
| Cheng et al. (2021) | Granger causality index | Time series | Chinese Shanghai A-share Index (SHAI), Korea Composite Stock Price Index (KOSPI) and Japan's Nikkei 225 Index (NIKKEI) | Finance | Time series |
| Chikahara & Fujino, (2018) | Granger causality index | Time series | Bivariate time series, Saccharomyces cerevisiae (yeast) cell cycle gene | Finance | Time series |
| Chivukula et al., (2018) | Granger causality | Time Series | Yahoo finance | Finance | Time series |
| Dastile et al., (2022) | Counterfactual | Tabular | German and HELOC | Financial Service | Explainability |
| Davis, (2021) | Counterfactual | Tabular | Security Prices, Funds flow data, Portfolio excess return | Finance | Financial economics |
| Geiger et al., (2013) | Granger causality index, Interventional | Time series, Stimulated data | Time series of 3 component | Finance | Time series |
| Gong et al., (2019) | Granger causality index | Graph data (as a Network) | 15 banking financial institutions, 2 insurance companies and 7 securities companies | Finance | Corporate Finance |
| Guelman & Guillén, (2022) | Counterfactual | Tabular | Canadian direct insurer. | Insurance | Automobile |
| Jacquier & Polson, (2012) | Bayesian Causal Network | Tabular | | Finance | Corporate Finance |
| Kleinberg et al., (2013) | Granger causality index | Time series | Stimulated financial time series and S&P 500 | Finance | Time series |

#Only finance domain is categorized further into subdomains.

**Table 6.** Summary of the surveyed papers (Contd.)

| References | Causal Inference approach | Input data type | Dataset | Domain# | Sub domain |
|---|---|---|---|---|---|
| Kolodiziev et al., (2014) | Statistical | Tabular | Ukrainian banking dataset | Banking | |
| Mc Grath et al., (2018)] | Counterfactual | Tabular | Home Equity Line of Credit (HELOC) | Financial services | Loan |
| Michail et al., (2019) | Counterfactual | Time series | Inflation estimates and bank lending growth of 3 European countries | Banking | |
| Ravivanpong et al., (2022) | Granger causality index | Time series | 1. Internal portfolio risk data 2. Deutsche Börse Public Dataset (PDS) | Finance | Financial Governance |
| Rudd, (2021) | Causal treatment | Tabular | Accounts of local members of finance company | Financial Services | Application (Churn) |
| Shah et al., (2019) | Counterfactual | Tabular | Retail business | Financial Services | Application (Churn) |
| Tsapeli et al., (2017) | Granger causality index | Time series | 1. Twitter sentiment analysis 2. NASDAQ | Finance | Behavioural Finance |
| Wang et al., (2023) | Counterfactual | Tabular | Quarterly Financial accounting statements from 332 companies | Finance | Corporate finance |
| Wu et al., (2021) | Pattern causality (PC) theory, cross-convergent mapping (CCM) theory | Time series | Morgan Stanley Capital International stock index | Finance | Time series |
| Yang et al., (2020) | Counterfactual | Financial Text | M&A dataset collected from Zephyr | Financial services | Explainability |

#Only finance domain is categorized further into subdomains.

Overall, the following are the research gaps identified in the literature:

- Almost all the research works presented in the survey paper employed causal inference under certain assumptions and conditions. Hence, achieving generalizability using these techniques is an important research direction.
- By employing causal inference, the performance of the employed model could be studied under adversary situations. This indeed helps in evaluating the risk or warning and helps in designing early warning systems, unlike traditional warning systems.
- Studies such as Kolodiziev (2014), Davis (2023), etc. employed only a few technical indicators while addressing the respective problems. Hence, including more technical indicators and studying the treatment is highly essential and could be a potential research direction.
- All the extant research works need to be designed in such a way that they could be applicable to the 5 Vs of Big Data, i.e., Velocity, Veracity, Volume, Value, and Variety. Hence, the scalability aspect needs to be considered, which could attract industry practitioners along with researchers.
- When applying to time-series related problems, the following could be potentially missing in the extant works:
    - Handling rare events occupies utmost importance now, especially when employing causal inference. These rare events could embody the critical events which are need to be analyzed and handled in a unique way.
    - Studying augmented data collected from various cross-sectional teams and studying various treatments is also a potential future direction.
    - Exploring time-variant and time-invariants and designing unique treatments is also highly essential.
    - Addressing unknown time delays also needs to be considered while employing various treatments.
- Improving the textual-based explanations, which are indeed important to understand the effects of counterfactuals. This indeed improves the quality of the explanations and makes it a better domain-agnostic model explanation.
- Handling the trade-off between sparsity and accuracy while employing causal inference also needs to be addressed. Further, employing XAI is also quite important to provide a better explanation of treatments.
- Designing causal analysis-enabled data science/AI/ML models for solving banking use cases such as churn prediction, fraud detection, stock market prediction, portfolio analysis at large scale, etc., could be the potential future directions.
- Selection bias is often the issue faced in the decision-making stage in every field. Hence, causal inference could be employed at that stage to study the impact of selection bias over results.

# 6. Tools and Software of Causal Inference

Some of the tools and software used for causal estimation and useful for BFSI is provided as summary in Table 7 and its explanation in subsequent paragraphs.

DoWhy (Sharma & Kiciman, 2020) is a comprehensive tool that is highly regarded in causal inference, developed using Python. DoWhy encompasses four main tasks: modeling the causal problem using a causal graph, identifying the desired causal estimand, estimating the causal effect, and validating the results obtained. The tool currently includes several identification strategies, such as the front-door criterion, and instrumental variables. Additionally, DoWhy seamlessly integrates with a broad range of machine learning-based estimators from EconML (Battocchi et al., 2023). EconML models enable users to conveniently choose the most suitable model for their specific query.

CausalML (Chen et al., 2020) is a Python package that offers a range of machine learning-based uplift modeling and causal inference methods, built on the latest research. It allows estimation of treatment effects at an individual level, enabling personalized recommendations and optimization. It supports uplift modeling, which measures the incremental impact of treatments on individual behavior using experimental data. For example, if a company wants to choose between multiple product lines to up-sell or cross-sell to its customers, CausalML can serve as a recommendation engine, identifying products that are likely to yield the greatest expected improvement for each user.

Propensity score matching, and doubly robust estimators are available in other tools such as Matching (Sekhon, 2011), CausalGAM (Glynn & Quinn, 2017), and ipw (Van Der Wal & Geskus, 2011). Doubly robust estimators can also be found in CausalGAM, an R library that offers both standard estimators and the AIPW estimator. The ipw package implements inverse probability of treatment weighting for both time-fixed and time-varying frameworks.

**Table 7**. Overview of Causal Inference estimation tools (Adapted from Nogueira et al.,2021)

| Software | References | Language | Estimands | Nature of UnConfoundedness | | | | No unconfoundedness | |
|---|---|---|---|---|---|---|---|---|---|
| | | | | Matching | Weighting | ML-based | DR | IV | Front door |
| **doWhy** | Sharma et al., (2019) | Python | ATE, LATE, ITE, CATE | ✓ | ✓ | ✓ | ✓ | ✓ | ✓ |
| **econML** | Battocchi et al., (2023) | Python | ATE, LATE, ITE CATE | | | ✓ | | ✓ | |
| **Matching** | Sekhon, (2011) | R | ATE, ATT, CATE | ✓ | | | | | |
| **CausalML** | Chen et al., (2020) | Python | ATE, LATE, ITE CATE | | | ✓ | | ✓ | |
| **Causal GAM** | Glynn & Quinn, (2017) | R | ATE, ATT | | ✓ | | ✓ | | |
| **ipw** | van der Wal & Geskus, (2011) | R | ATE, ATT, CATE | | ✓ | | ✓ | | |

# 7. Implications for Managers and Researchers

In this section, we discuss the implications of employing CI for both managers and researchers. First, we discuss implications from the researchers' perspective followed by managerial perspective.

## 7.1 Implications for academic researchers

In this survey, we presented a comprehensive overview of causal inference and the types of methods related to various applications in the field of BFSI. We followed very easy-to-follow procedures and presented a comprehensive overview of the research papers. For ease of understanding, we segregated the articles as per the domains and their applicability. This surely attracts the researchers not only limited to finance, economics but also from various fields such as econometrics, computer science, etc. This survey embodies the fact that causal inference is employed in the field of BFSI sector since the early 1990s. However, a large amount of research happened in the siloed manner, which resulted in slow progress. The recent surge in the number of publications indicates that a lot of advancements are about to happen in the coming years in this field.

Further, the other important implications for academic researchers are listed below:

In our survey paper, we considered finance and its subdomains such as accounting, corporate finance, corporate governance. Further, we also explore explainability aspect which is becoming a prominent and one of the critical aspects in recent times especially in the domains such as in financial services, which opens the development of causal inference-based methods. With this survey paper, we are trying to intricate the relationships between causal inference within the financial landscape. In doing so, the decision-making aspects will be further improved leveraging the employment of multifaceted aspects with better sustainable and resilient financial ecosystem.

It is observed to have a plethora of opportunities especially from an application point of view, within the financial landscape. Notably, the researchers and scientists can look at the various financial service-based use cases such as credit lending, credit rating and churn modelling in insurance and banking. Now we will discuss the case studies in detail. In credit rating, the risk assessment stands out to be the critical aspect. Hence, there is a requirement of sophisticated methodologies which could lead to better investment decisions with a reduce in the risk involved. In churn modeling, identifying the churned out customers indeed boosts the economic growth of an organization. Further, it helps the organization to revisit their policies and helps in building the association between the current customers and the respective organizations. In this way, by focusing more on these financial driven use cases, academicians, researchers and scientists can contribute in the betterment of the society.

Furthermore, incorporating the extended CRISP-DM to include explainability and causal inference in their proposed methodologies indeed increases the robustness, and trustworthiness of the proposed methodology. By adopting this extended framework, the researchers, scientists can provide valuable insights to understand how the predictions are performed by the deployed machine learning / deep learning models, what are the influencing variables or features, eradicating methods, customer centric variables etc., will increase the revenue growth of a financial organization. The transparency is one thing which will foster the greater understanding and acceptance of the results and will increase the confidence over the machine learning models. Additionally, employing causal inference will give cause and effect relationships which are more reliable and helps in making proper decision-making.

Through this survey, we want to project the observed consistent trend in the improvement of the methodologies without concerning over the privacy aspects in their studies. This is indeed essential and a substantially valuable aspect which will increase the confidence of the customers over the organizations. In due process, it will increase the customer satisfaction and binds the customer organization relationship well. Therefore, the inclusion of the privacy aspect is also an essential and interesting future research direction.

## 7.2 Implications for managers

Our study is not only limited to attract academic researchers, but also targets the practitioners of the BFSI area. We presented a comprehensive review of the research papers. For the ease of understanding, we segregated the articles as per the domains and their applicability. By studying the survey paper, practitioners and managers will get a holistic overview of the importance of incorporating causal inference into their data science/AI/ML initiatives. Further, we presented the extended CRISP-DM, which will signify the importance of the Causal inference into their present CRISP-DM workflow. Other important implications for managers are follows:

Incorporating CI and XAI supports post-hoc explaniability of the models built to solve various operational and business problems brings substantial benefits. By utilizing CI, increasing confidence and trust over the deployed models. This will also help the stake holders to gain a clear interpretable picture of the model's behavior. Additionally, it will provide transparent insights into how th emodel arrives at the specific decisions. Further, this will makes the validating, fine-tuning processes easier than the earlier methods. Further, it presents a trustworthy picture of what-if scenarios and necessary decision-making policies.

After employing the modified CRISP-DM framework, helps the managers study to the scenarios upfront, which would improve the business outcomes. By gaining more insights through studying causal and effect relationships between different variables, which are potentially effecting the outcomes of the business process helps in inclusion the new policies as per the trend. Having this said, will improve the decision-making policies and the organization will walk towards the customer centric approaches. This kind of proactive mechanisms enables the managers to optimize their business strategies and maximize the potential outcomes.

In addition to the above business perspectives, it also helps the managers to understand whether the deployed model are in compliance with the regulatory requirements or not. If they are not complying as per the regulatory then there may be lot of issues such as privacy leakage, improbable decision making, bias over a few sections of the society etc., can be eradicated. This will give a transparent view of the decision-making process and helps the managers to understand the highly influencing variables. Further, it will also help avoid the financial penalties, legal repercussions and reputational damage.

Causal inference helps practitioners to study adversary situations early on, thereby leveraging them to design a future-proof plan. The adversary is a kind of unwanted scenario where the attacker probes the system and gathers the critical information by sending numerous queries. This will increase the false positives which creates huge financial and reputational losses to the organization. Hence, studying them earlier by employing CI and XAI will eradicate this problem and prepares the organization to tackles these kind of situations in a better way. Further, it will make the model more resilient and robust towards these kind of uncertainties.

Furthermore, we presented future research directions, which are open research problems at the fag-end of our review paper. The researchers, practitioners, academicians will find them extremely useful.

## 8. Conclusions

In this paper, we comprehensively surveyed several works highlighting the usefulness of causal inference in solving diverse problems in the BFSI area. Further, especially during the recent past five years, there is an dramatic rise in the development of causal inference-based methodologies, studying the counterfactuals, etc., to design a robust mechanism. However, in the banking and insurance sectors, causal inference is still in its infancy, and thus more research is warranted in this niche area to turn it into a viable method. In time series, a lot of research has been conducted due to the use of Granger causality.

# 9. Future research directions

After surveying these papers, we identified the potential open research problems as follows:

**Counterfactual optimization formulated as Multi Objective Optimization (MOO) problem:** To accomplish this objective, one can employ MOO algorithms such as Non dominated sorting optimization algorithm (NSGA-II), Multi-objective optimization algorithm with decomposition (MOEA/D), Non-dominated sorting particle swarm optimization (NSPSO) etc. In the literature, genetic programming has been extensively used for optimizing the counterfactuals, where proximity and diversity are the objective functions (ref).

**Bankruptcy prediction using Causal analysis:** one can potentially investigate in what situation a bank goes bankrupt, by incorporating causal analysis as prescriptive analytics tool.

**Cyber-attack in Intrusion Detection:** Generated counterfactuals can be used as evasion samples to evaluate how the deployed intrusion detection algorithms will perform in adverse situations.

**Churn Analysis using Intervention and Optimization of Counterfactuals:** Current machine learning models can only discriminate whether the customer will churn out or not. However, what must be done in-order to retain loyal customers and win back the churned customers is recommended by causal analysis.

**Causal Analysis in ATMs placement:** Causal analysis can be further applied to the combinatorial optimization problems such as suitable placement of ATMs, suitable policy recommendations, etc. which have operational significance.

**User behavior analysis:** CI also has a potential role in obtaining the correlation between user behavior patterns and adverse users, which will lead to an increase in the potential candidates etc.

**Causal Inference in Knowledge based graphs:** Knowledge based graphs possess critical information pertaining to the domain where they are applied. Hence, perturbing it could yield more insights that could potentially improve the business.

**Extension of counterfactuals to other insurance problems:** Further, one can extend these studies not only in healthcare-based insurance but also in many other insurances such as travel insurance, life insurance, and fire insurance. Currently, counterfactual studies have been carried out in automobile insurance.

**Studying the impact of missing values:** Often real-world datasets have missing values. However, it adversely influences decision making, especially in the finance and healthcare domains. Hence, studying the adverse effects caused by missing values and their effect on business strategy is also a potential problem.